\pdfoutput=1

\documentclass[11pt]{article}

\usepackage[preprint]{acl}
\usepackage{colortbl}
\usepackage{tabularx}
\usepackage{arydshln}
\usepackage{times}
\usepackage{latexsym}
\usepackage{adjustbox}
\usepackage{caption}
\usepackage{subcaption}
\usepackage{booktabs}
\usepackage{graphicx}
\usepackage{array}
\usepackage{multirow}
\usepackage{multicol}
\usepackage{booktabs}
\usepackage{pgfplots}
\usepackage{url}
\usepackage{enumitem}
\usepackage{amsmath}
\usepackage{amssymb}
\usepackage{algpseudocode}
\usepackage{algorithm}
\usepackage{adjustbox}
\usepackage{fontawesome5}
\usetikzlibrary{arrows}
\usetikzlibrary{shapes.arrows}
\usepackage{bbding}
\usepackage{pifont}

\newcommand{\cmark}{\ding{51}}%
\newcommand{\xmark}{\ding{55}}%

\algnewcommand\algorithmicforeach{\textbf{for each}}
\algdef{S}[FOR]{ForEach}[1]{\algorithmicforeach\ #1\ \algorithmicdo}

\usepackage{tikz}
\usetikzlibrary{shapes,arrows,positioning}
\usepgfplotslibrary{groupplots}
\usetikzlibrary{pgfplots.groupplots}
\usetikzlibrary{plotmarks}
\usetikzlibrary{calc}
\usepgfplotslibrary{external}
\usetikzlibrary{positioning}

\pgfdeclarelayer{foreground}
\pgfsetlayers{main,foreground}

\usetikzlibrary{patterns}
\pgfplotsset{compat=1.16}

\definecolor{lightblue}{RGB}{178,178,255}
\definecolor{darkgreen}{RGB}{51,153,51}
\definecolor{overlapcl}{RGB}{153,184,194}
\definecolor{forestgreen}{rgb}{0.13, 0.55, 0.13}
\colorlet{lightgrey}{white!90!black}
\colorlet{mixedcolor}{darkgreen!50!lightblue}

\usepackage{textcomp}
\usepackage[T1]{fontenc}

\usepackage[utf8]{inputenc}

\usepackage{microtype}

\usepackage{inconsolata}

\title{Anchor-based Large Language Models}

\author{Jianhui Pang$^{1}$\thanks{Work was done when Jianhui Pang and Fanghua Ye were interning at Tencent AI Lab.}~~~~~Fanghua Ye$^{2}$\footnotemark[1]~~~~~\bf Derek Fai Wong$^{1}$\footnotemark[2]~~~~~\bf Xin He$^{3}$\\ \bf Wanshun Chen$^{3}$~~~~~\bf Longyue Wang$^{3}$\thanks{Corresponding Authors.}\\$^1$University of Macau~~~~~~$^2$University College London~~~~~$^3$Tencent AI Lab\\\normalsize nlp2ct.pangjh3@gmail.com, fanghua.ye.19@ucl.ac.uk, derekfw@um.edu.mo \\ \normalsize \{shaynechen, kleinhe, vinnylywang\}@tencent.com}

\begin{document}
\maketitle
\begin{abstract}

Large language models (LLMs) predominantly employ decoder-only transformer architectures, necessitating the retention of keys/values information for historical tokens to provide contextual information and avoid redundant computation. However, the substantial size and parameter volume of these LLMs require massive GPU memory. This memory demand increases with the length of the input text, leading to an urgent need for more efficient methods of information storage and processing. This study introduces Anchor-based LLMs (AnLLMs), which utilize an innovative anchor-based self-attention network (AnSAN) and also an anchor-based inference strategy. This approach enables LLMs to compress sequence information into an anchor token, reducing the keys/values cache and enhancing inference efficiency.
Experiments on question-answering benchmarks reveal that AnLLMs maintain similar accuracy levels while achieving up to 99\% keys/values cache reduction and up to 3.5 times faster inference.
Despite a minor compromise in accuracy, the substantial enhancements of AnLLMs employing the AnSAN technique in resource utilization and computational efficiency underscore their potential for practical LLM applications.\footnote{Our code and models are publicly available at: \url{https://github.com/pangjh3/AnLLM}.}

\end{abstract}



\section{Introduction}

Large language models (LLMs) primarily utilize decoder-only transformer architectures, which necessitate caching keys/values information for historical tokens during the auto-regressive inference to supply contextual information and avoid redundant computation \cite{wei2022emergent,touvron2023llama,openai2023gpt4,touvron2023llama2}. However, due to their immense size and high parameter count, a considerable amount of GPU memory is required for loading. 
Furthermore, as the length of input text grows, storing keys/values caches requires more and more GPU memory, as evidenced in in-context learning, complex instructions, and extended conversations \cite{dong2022survey,jiang2023longllmlingua,wang-etal-2023-label}, which is not conducive to scenarios with limited computational resources.
An alternative approach entails recalculating these extensive inputs, which, however, results in increased time overhead. Therefore, this study aims to \textit{reduce the storage demand for keys/values caches during the inference phase of LLMs, improving the memory efficiency and, consequently, accelerating the inference speed as well}.

\begin{figure}
    \centering
\resizebox{0.45\textwidth}{!}{
\begin{tikzpicture}
    \begin{axis}[
        ybar,
        legend columns=1,legend style={font=\small},
        legend pos=north west,
        enlargelimits=0.15,
        ylabel={\bf {\color{forestgreen}Keys/Values Caches}},
        xtick={0,1,2},
        xticklabels={OBQA, PIQA, BoolQ},
        ymajorgrids=true,
        x tick label style={yshift=0.15cm},
        grid style=dashed,
        nodes near coords,
        nodes near coords align={vertical},
        width=0.45\textwidth,
        height=2.5in,
        tick style={draw=none},
        ]
    \addplot [pattern=grid, pattern color=forestgreen, thick,line width=1pt] coordinates { (0,89) (1,262) (2,804)};
    \addplot [fill=black, thick,line width=1pt] coordinates {(0,5) (1,5) (2,5)};
    
    \legend{Text lengths, Our caches}
    
    \end{axis}

\begin{axis}[
    axis y line*=right,
    ylabel={\bf {\color{blue}Acceleration Ratio}},
    enlargelimits=0.15,
    xtick={0,1,2},
    xticklabels={OBQA, PIQA, BoolQ},
    ytick={1,1.5,3.5},
    yticklabels={{\small $\times$}1.0, {\small $\times$}1.5, {\small $\times$}3.5},
    grid style=dashed,
    height=2.5in,
    width=0.45\textwidth,
    nodes near coords,
    point meta=explicit symbolic
]

\addplot[
    color=blue,
    mark=triangle,
    mark options={scale=1.5}
    ]
    coordinates {
    (0,1.0)(1,1.5)(2,3.5)
    };

\end{axis}

\end{tikzpicture}
}

\caption{Keys/Values Caches and Inference Acceleration Ratio of Ours in OBQA, PIQA, and BoolQ Tasks with Five-Shot Demonstrations. The bars indicate Keys/Values cache and text length, while the curve shows the Inference Acceleration Ratio. As text length increases, our method achieves up to 99\% reduction in Keys/Values Caches compared to traditional approaches. Moreover, caching prefix texts enhances inference efficiency by 3.5 times over non-caching methods.}


    \label{fig:startfig}
\end{figure}
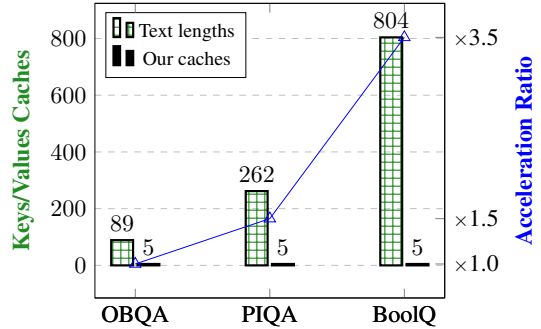

In a recent study, \newcite{wang-etal-2023-label} demonstrate that label words in prefix demonstrations can act as anchors during inference, providing an effective context compression approach for improving inference efficiency in in-context learning. However, in practical applications, not all prefix inputs or demonstrations contain label words suitable for compressing information, making the reliance on label words a less universal approach for text information compression.
Additionally, \newcite{pang2024salute} observe that LLMs tend to attend to only a few, yet consistent, prefix tokens during inference. However, the specific tokens utilized are often unpredictable and uncontrollable. These observations raise an intriguing question: \emph{do natural language texts contain anchor points that compress the overall semantic information of sequences?} 
In this context, previous studies on sequence embeddings have shown that the hidden state of a special token in neural network models can encapsulate semantic information \cite{baudis-etal-2016-joint, devlin2018bert}. 
Furthermore, contemporary LLMs typically utilize the causal self-attention mechanism during both training and inference phases \cite{touvron2023llama,touvron2023llama2}, attending on each prior token. 
This suggests that the final token in a sequence may be better suited to serve as a natural information compression point compared to other tokens, as they cannot observe future tokens. 
Therefore, a methodical approach that identifies and exploits sequence anchor tokens in a dependable and controllable manner is essential for compressing sequence information, effectively reducing keys/values caches, and improving inference efficiency for LLMs.

To this end, we propose novel \textbf{An}chor-based \textbf{L}arge \textbf{L}anguage \textbf{M}odels (AnLLMs), equipped with an innovative anchor-based self-attention network (AnSAN) and an anchor-based inference strategy. 
The AnSAN is designed to compel the models to compress sequence information into the anchor token (the last token in our implementation) during the training process, with the aid of anchor-based attention masks.
During inference, the anchor-based inference strategy retains the keys/values caches of anchor tokens, which have aggregated the entire sequence information, and discards those of non-anchor tokens, thereby reducing memory demands.
Specifically, the anchor-based attention masks for AnSAN serve two objectives: 1) to ensure anchor tokens attend exclusively to tokens within the same sequence, preventing attention to other sequences, and 2) to direct non-anchor tokens' attention to previous sequence anchors, blocking the other non-anchor tokens from previous sequences.
It is noteworthy that the technique of anchor-based attention bears similarities to the principles underlying sparse attention \cite{child2019generating}. However, unlike the existing research that employs sparse attention to extend the context length of LLMs \cite{longlora,ratner-etal-2023-parallel}, our method focuses on continually pre-training the model to compress sequence information into the anchor token. 

In our implementation, we utilize the publicly available RedPajama datasets \cite{together2023redpajama} to continuously pre-train the open-source Llama2 models \cite{touvron2023llama2}, resulting in AnLLMs that incorporate our proposed anchor-based attention mechanism. 
Experimental results on question answering benchmarks, as depicted in Figure \ref{fig:startfig}, reveal that our method achieves up to a 99\% reduction in keys/values caches and up to a 3.5-fold increase in inference acceleration ratios, while maintaining comparable accuracy to the original model. Despite a minor decrease in accuracy (within 1.5\%), these findings underscore the significant improvements in computational efficiency and memory utilization offered by our method.

\section{Related Work}

Our research is inspired by the recent investigation into the understanding of in-context learning (ICL) within LLMs by \citet{wang-etal-2023-label}. In their study, the authors delve into the underlying mechanisms of ICL, emphasizing the influence of label words in demonstration examples on information flow. They reveal that these label words serve as anchors, wherein semantic information converges into these anchors during inference, subsequently directing the LLMs' final predictions.
Motivated by their findings, our objective is to extend this feature to natural language modeling by guiding sequence information compression into manually designed anchor tokens, rather than solely relying on label words. This is crucial because natural language texts may not always contain an explicit label.

The most relevant method to our approach in the existing literature is the learning to compress prompts with gist tokens \cite{mu2023learning}. 
Their approach centers around compressing task-specific prompts by fine-tuning the model using the proposed gist masking, thereby enforcing prompt compression. 
However, there are several crucial divergences between our study and theirs. 
Unlike their focus on compressing a task prompt, our objective lies in training the LLM to condense sequence information into the anchor tokens. 
Consequently, our approach can be universally applied to a range of tasks without requiring task-specific training, a feature not shared by gist tokens, as the anchor tokens are seamlessly incorporated into the model's language modeling. 
Furthermore, our anchor-based attention masks account for information compression within a sequence and information interaction between sequences, thus extending beyond the mere compression of task prompts.

On the other hand, Flash{A}ttention \cite{dao2022flashattention} and PagedAttention \cite{kwon2023efficient} both present memory-efficient attention mechanisms for LLMs. While they focus on optimizing attention computation and subdividing attention processing, 
our proposed method offers a distinct approach that specifically targets the compression of sequence information into anchor tokens, making it orthogonal to these existing works.



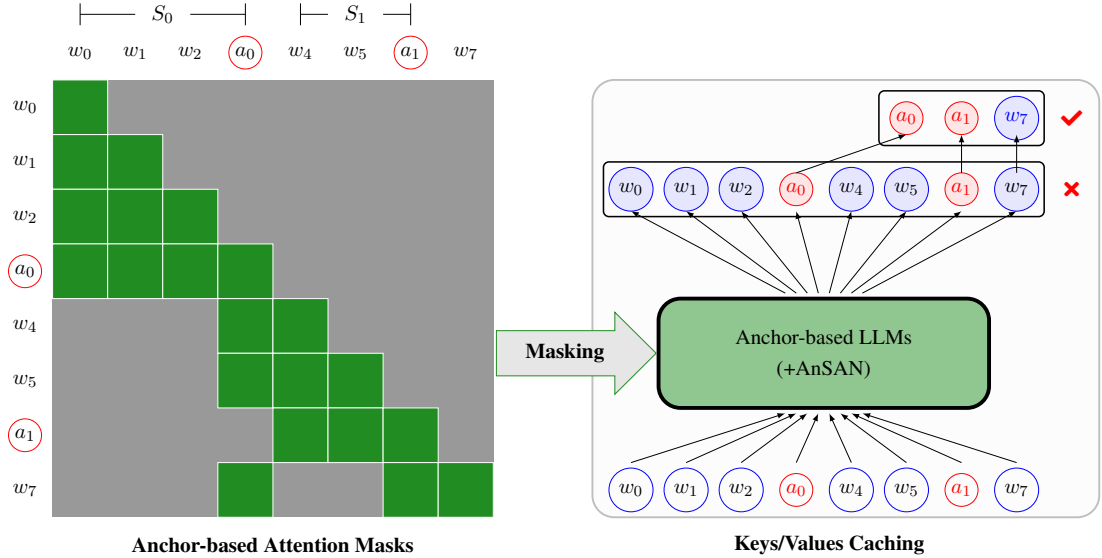
\begin{figure*}[ht]
    \centering
\begin{adjustbox}{width=0.9\textwidth}
\begin{tikzpicture}
\foreach \i in {0,1,...,7} {
    \foreach \j in {0,1,...,7} {
        \fill[gray!80] (\i,\j) rectangle +(1,1);
    }
}

\foreach \i in {0,1,2,4,5,7} {
    \node at (\i+0.5,+8.5) {$w_{\i}$};
}
\node at (3+0.5,+8.5) {$a_{0}$};
\node at (6+0.5,+8.5) {$a_{1}$};

\foreach \i in {0,2,3,5,6,7} {
    \pgfmathtruncatemacro{\inverted}{7 - \i}
    \node at (-0.5,\i+0.5) {$w_{\inverted}$};
}
\node at (-0.5,1+0.5) {$a_{1}$};
\node at (-0.5,4+0.5) {$a_{0}$};

\fill[forestgreen, draw=white, line width=0.5pt] (0,7) rectangle (1,8);
\fill[forestgreen, draw=white, line width=0.5pt] (0,6) rectangle (1,7);
\fill[forestgreen, draw=white, line width=0.5pt] (1,6) rectangle (2,7);
\fill[forestgreen, draw=white, line width=0.5pt] (0,5) rectangle (1,6);
\fill[forestgreen, draw=white, line width=0.5pt] (1,5) rectangle (2,6);
\fill[forestgreen, draw=white, line width=0.5pt] (2,5) rectangle (3,6);
\fill[forestgreen, draw=white, line width=0.5pt] (0,4) rectangle (1,5);
\fill[forestgreen, draw=white, line width=0.5pt] (1,4) rectangle (2,5);
\fill[forestgreen, draw=white, line width=0.5pt] (2,4) rectangle (3,5);
\fill[forestgreen, draw=white, line width=0.5pt] (3,4) rectangle (4,5);
\fill[forestgreen, draw=white, line width=0.5pt] (3,3) rectangle (4,4);
\fill[forestgreen, draw=white, line width=0.5pt] (4,3) rectangle (5,4);
\fill[forestgreen, draw=white, line width=0.5pt] (3,2) rectangle (4,3);
\fill[forestgreen, draw=white, line width=0.5pt] (4,2) rectangle (5,3);
\fill[forestgreen, draw=white, line width=0.5pt] (5,2) rectangle (6,3);
\fill[forestgreen, draw=white, line width=0.5pt] (4,1) rectangle (5,2);
\fill[forestgreen, draw=white, line width=0.5pt] (5,1) rectangle (6,2);
\fill[forestgreen, draw=white, line width=0.5pt] (6,1) rectangle (7,2);
\fill[forestgreen, draw=white, line width=0.5pt] (3,0) rectangle (4,1);
\fill[forestgreen, draw=white, line width=0.5pt] (6,0) rectangle (7,1);
\fill[forestgreen, draw=white, line width=0.5pt] (7,0) rectangle (8,1);

\draw (0.5, 9.2) -- (1.5, 9.2);
\draw (0.5, 9.0) -- (0.5, 9.4);
\node at (2.0, 9.2) {$S_{0}$};
\draw (2.5, 9.2) -- (3.5, 9.2);
\draw (3.5, 9.0) -- (3.5, 9.4);

\draw (4.5, 9.2) -- (5.0, 9.2);
\draw (4.5, 9.0) -- (4.5, 9.4);
\node at (5.5, 9.2) {$S_{1}$};
\draw (6.0, 9.2) -- (6.5, 9.2);
\draw (6.5, 9.0) -- (6.5, 9.4);


\draw[red] (3.5, 8.5) circle (0.3cm);
\draw[red] (6.5, 8.5) circle (0.3cm);
\draw[red] (-0.5, 1.5) circle (0.3cm);
\draw[red] (-0.5, 4.5) circle (0.3cm);

\node at (4, -0.5) {\bf Anchor-based Attention Masks};


\begin{scope}[xshift=10cm]

\node at (4.1, -0.5) {\bf Keys/Values Caching};

\draw[gray!50, fill=gray!2, rounded corners=4mm, line width=1pt] (-0.2,0.0) rectangle (9.0, 8.0);

    \draw[blue] (0.5, 0.5) circle (0.4cm) node {\textcolor{black}{$w_{0}$}};
    \draw[blue] (1.5, 0.5) circle (0.4cm) node {\textcolor{black}{$w_{1}$}};
    \draw[blue] (2.5, 0.5) circle (0.4cm) node {\textcolor{black}{$w_{2}$}};
    \draw[red] (3.5, 0.5) circle (0.3cm) node {$a_{0}$};
    \draw[blue] (4.5, 0.5) circle (0.4cm) node {\textcolor{black}{$w_{4}$}};
    \draw[blue] (5.5, 0.5) circle (0.4cm) node {\textcolor{black}{$w_{5}$}};
    \draw[red] (6.5, 0.5) circle (0.3cm) node {$a_{1}$};
    \draw[blue] (7.5, 0.5) circle (0.4cm) node {\textcolor{black}{$w_{7}$}};
    
    \draw[rounded corners=4mm, thick, fill=forestgreen!50,line width=2pt] (1,2) rectangle (7, 4);
    \node at (4, 3.3) {Anchor-based LLMs};
    \node at (4, 2.7) {(+AnSAN)};

    \node[single arrow, draw=forestgreen, minimum height=2.9cm, minimum width=1.5cm, fill=gray!20] at (-0.7, 3) {\bf Masking};

    \draw[-latex] (0.5, 1.0) -- (3.3, 1.9);
    \draw[-latex] (1.5, 1.0) -- (3.5, 1.9);
    \draw[-latex] (2.5, 1.0) -- (3.7, 1.9);
    \draw[-latex] (3.5, 1.0) -- (3.9, 1.9);
    \draw[-latex] (4.5, 1.0) -- (4.1, 1.9);
    \draw[-latex] (5.5, 1.0) -- (4.3, 1.9);
    \draw[-latex] (6.5, 1.0) -- (4.5, 1.9);
    \draw[-latex] (7.5, 1.0) -- (4.7, 1.9);

    \draw[-latex] (3.3, 4.1) -- (0.5, 5.6);
    \draw[-latex] (3.5, 4.1) -- (1.5, 5.6);
    \draw[-latex] (3.7, 4.1) -- (2.5, 5.6);
    \draw[-latex] (3.9, 4.1) -- (3.5, 5.6);
    \draw[-latex] (4.1, 4.1) -- (4.5, 5.6);
    \draw[-latex] (4.3, 4.1) -- (5.5, 5.6);
    \draw[-latex] (4.5, 4.1) -- (6.5, 5.6);
    \draw[-latex] (4.7, 4.1) -- (7.5, 5.6);

    \draw[blue, fill=blue!10] (0.5, 6.0) circle (0.4cm) node {\textcolor{black}{$w_{0}$}};
    \draw[blue, fill=blue!10] (1.5, 6.0) circle (0.4cm) node {\textcolor{black}{$w_{1}$}};
    \draw[blue, fill=blue!10] (2.5, 6.0) circle (0.4cm) node {\textcolor{black}{$w_{2}$}};
    \draw[red, fill=red!10] (3.5, 6.0) circle (0.3cm) node {$a_{0}$};
    \draw[blue, fill=blue!10] (4.5, 6.0) circle (0.4cm) node {\textcolor{black}{$w_{4}$}};
    \draw[blue, fill=blue!10] (5.5, 6.0) circle (0.4cm) node {\textcolor{black}{$w_{5}$}};
    \draw[red, fill=red!10] (6.5, 6.0) circle (0.3cm) node {$a_{1}$};
    \draw[blue, fill=blue!10] (7.5, 6.0) circle (0.4cm) node {\textcolor{black}{$w_{7}$}};
    \draw[rounded corners=1mm, thick] (0,5.5) rectangle (8, 6.5);
    \node at (8.5, 6.0) {\textcolor{red}{\faTimes}};

    \draw[red, fill=red!10] (5.5, 7.3) circle (0.3cm) node {$a_{0}$};
    \draw[red, fill=red!10] (6.5, 7.3) circle (0.3cm) node {$a_{1}$};
    \draw[blue, fill=blue!10] (7.5, 7.3) circle (0.4cm) node {$w_{7}$};
    \draw[rounded corners=1mm, thick] (5,6.8) rectangle (8,7.8);
    \node at (8.5, 7.3) {\textcolor{red}{\faCheck}};

    \draw[-latex] (3.5,6.3) -- (5.5,7.0);
    \draw[-latex] (6.5,6.3) -- (6.5,7.0);
    \draw[-latex] (7.5,6.3) -- (7.5,7.0);
    

\end{scope}

\end{tikzpicture}
\end{adjustbox}
\caption{Anchor-based Attention Masking and Efficient Caching in Anchor-based LLMs. On the left, the gray and green squares represent the masking and unmasking operations respectively, with the circled ``a'' symbols denoting the anchor tokens. On the right, the shaded circles depict keys/values caches. By employing anchor-based attention masking during training, we compel the model to compress sequence information into the anchor tokens. On the right, during inference, with the AnSAN technique, AnLLMs compress information into the anchor tokens and discard the previous remaining keys/values caches, thereby facilitating an efficient caching mechanism.}


\label{fig:mainfigure}
\end{figure*}

\section{Anchor-based Large Language Models}
\label{sec:allm}

\subsection{Background}
\label{sec:background}


\paragraph{Transformers.} LLMs are primarily realized as decoder-only transformers \cite{vaswani2023attention,touvron2023llama,touvron2023llama2}, incorporating an input embedding layer and multiple decoder layers. Each layer contains a self-attention network and a feed-forward network with normalization modules. Crucially, causal attention masks are employed, allowing tokens to attend only to preceding ones.

\paragraph{Self-Attention Networks.} Typically for decoder-only LLMs like Llama2 \cite{touvron2023llama2}, self-attention networks (SANs) map queries $Q$, keys $K$, and values $V$ into an output, as delineated in the following equations,
\begin{gather}
\small
    \text{SAN}(Q,K,V) = \text{Softmax}(Q,K)V, \label{eq:sans1} \\
\small
    \text{Softmax}(Q,K)_{i,j} = \frac{M_{i,j}\text{exp}(Q_{i}K_{j}^{T})}{\Sigma_{k}M_{i,k}\text{exp}(Q_{i}K^{T}_{k})}, \label{eq:sans2} \\ 
\small    M_{i,j}=  
    \begin{cases}
    1, & \text{if }i \geq j \\
    0, & \text{else } 
  \end{cases}, \label{eq:sans3}
\end{gather}
\noindent where $M$ denotes an $L\times L$ masking matrix, facilitating the current $i$-th token to attend to only preceding tokens whilst disregarding subsequent tokens during the training and inference phases.

\paragraph{Keys/Values Caches.} In the application of LLMs, the keys/values caches increase with lengthy prefix texts and continuously generated tokens during the inference phase, such as in question-answering \cite{saad2023pdftriage}, text summarization \cite{basyal2023text}, and machine translation \cite{pang2024salute}.
The key and value matrices associated with tokens of prefix inputs are cached to avoid recomputation and expedite subsequent token prediction \cite{radford2019language}.
Additionally, the model generates the output token-by-token in the real-time inference process, which requires more cache memory to store the newly generated sequence.
Therefore, addressing the challenges arising from the ever-expanding texts is crucial for enhancing the efficiency of LLM inference.

\subsection{Anchor-based Self-Attention Networks}
\label{sec:AbSAN}


Given an input text with $n$ consecutive sequences, $P=\{S_{1},S_{2},...,S_{n}\}$, their associated anchor tokens are the last tokens that represented as $A=\{a_{1},a_{2},...,a_{n}\}$.
The primary objective of AnSAN is to encapsulate the information of a sequence into its anchor token, with the anchor hidden states representing the comprehensive semantic information.
In this manner, an AnLLM equipped with AnSAN generates subsequent tokens based on the keys/values caches of preceding tokens within the current sequence and the keys/values caches of anchor tokens from previous sequences.


\paragraph{Anchor-based Attention Masks.} 
To accomplish this, we devise anchor-based attention masks, as illustrated in Figure~\ref{fig:mainfigure}. 
Assuming that the current token in the sequence is a non-anchor token, we allow attention towards previous non-anchor tokens within the same sequence and anchor tokens from preceding sequences, while blocking attention towards non-anchor tokens from previous sequences. 
This approach ensures that non-anchor tokens can only access information from anchor tokens in previous sequences and the current sequence's information. 
Conversely, when the current token is an anchor token, which is the last token in the sequence, we exclusively permit its attention towards previous non-anchor tokens within the same sequence, blocking all other attention. 
This constraint forces the anchor token to aggregate information solely from its current sequence. 
Consequently, we replace Eq.~\eqref{eq:sans3} with anchor-based attention masks in Eq.~\eqref{eq:anchormask} to determine the mask of the $i$-th token in the input text concerning the $j$-th token (assuming that the $i$-th token belongs to the $k$-th sequence). 
\begin{gather}
\small
    M_{i,j}=  
    \begin{cases}
    0, & \text{if }((w_{i}, w_{j})\notin A) \land (w_{j} \in S_{\leq k-1})  \\
    0, & \text{else if }(w_{i} \in A) \land (w_{j} \in S_{\leq k-1})  \\
    1, & \text{else if }i \geq j  \\
    0, & \text{else} \\
  \end{cases}, \label{eq:anchormask}
\end{gather}
\noindent where $S_{\leq k-1}$ represents previous $k$\textminus$1$ sequences. The number $0$ denotes blocking attention, whereas the number $1$ indicates the opposite.



\paragraph{Anchor Token Selection.}

By implementing the AnSAN mechanism for training LLMs, we can compel the model to compress sequence information into the anchor token and generate new tokens based on the anchor token information from previous sequences and non-anchor token information from the current sequence.

The challenge now lies in selecting an appropriate anchor token. In our experiments, we propose two implementation methods: one using the endpoint as the anchor token, and the other appending a new token specifically as the anchor token.


\subsection{Anchor-based Inference}

By training the model to compress information into the anchor token of a natural language sequence, we can optimize the inference process by modifying the keys/values caching mechanism. Specifically, during inference, upon encountering an anchor token that condenses the comprehensive semantic information of preceding tokens in the current sequence, the model can reduce the keys/values caches by deleting the caches of non-anchor tokens within that sequence.

\begin{algorithm}[ht]
    \caption{Anchor-based Inference}\label{algo:anchorbasedinfer}
    \small
    \begin{algorithmic}[1]
        \Require Anchor-based LLM $\Theta$, prefix text $P$ with anchor tokens, keys/values cache list $\mathcal{C}$, predicted token $w_{new}$;
        \Ensure Generated text $\mathcal{T}$;
        \Function{Reduction}{$\mathcal{C}$}
            \State $j \leftarrow$ last anchor index in $\mathcal{C}$;
            \State $\mathcal{C} \leftarrow \{c \in \mathcal{C} \,|\, \text{index}(c) \geq j \, \text{or} \, c \,\text{is anchor}\}$;
            \State \Return $\mathcal{C}$.
        \EndFunction
        \State Initialize $\mathcal{T}, \mathcal{C}$ as empty lists;
        \State $\mathcal{M} \leftarrow$ GetMasks($P, \mathcal{C}$) using Eq.~\eqref{eq:anchormask};
        \State Update $w_{new}$, $\mathcal{C}$ using Forward($P$;$\mathcal{M}, \mathcal{C}, \Theta$);
        \State Append $w_{new}$ to $\mathcal{T}$;
        \State $\mathcal{C} \leftarrow$ Reduction($\mathcal{C}$);
        \While{$w_{new}$ is not [eos]}
            \State $\mathcal{M} \leftarrow$ GetMasks($w_{new}$, $\mathcal{C}$) using Eq.~\eqref{eq:anchormask};
            \State Update $w_{new}$, $\mathcal{C}$ using Forward($w_{new}$;$\mathcal{M}, \mathcal{C}, \Theta$);
            \State Append $w_{new}$ to $\mathcal{T}$;
            \If {$w_{new}$ is the anchor token}
                \State $\mathcal{C} \leftarrow$ Reduction($\mathcal{C}$);
            \EndIf
        \EndWhile
        \State \Return {$\mathcal{T}$}.
    \end{algorithmic}
\end{algorithm}


We introduce the inference method in Algorithm~\ref{algo:anchorbasedinfer}. The function ``R{\small{EDUCTION}}'' in Line 1 is utilized to remove keys/values caches when the model processes prefix texts in Line 10 or generates an anchor token during the prediction of the next token in Line 16. 
This approach aims to reduce the keys/values caches for both prefix tokens and generated outputs during real-time inference.



\input{figures/trainingloss_and_ppl}

\section{Experimental Setup}


In this section, we first detail AnLLM's implementation, then outline the training procedure and model perplexity. Finally, we introduce the evaluation datasets and metrics.

\subsection{Our Implementation}

Llama2-7b \cite{touvron2023llama2} is adopted as the base model in our experiments, which is an open-source and English-centric LLM.
In accordance with the principles outlined in Section~\ref{sec:allm}, we present our implementations here. The crux is to identify which tokens in a sequence can be considered anchor tokens. 
In light of this, we describe two implementation strategies: one employs the endpoints directly, and the other involves appending a new token at the end of a sequence to serve as the anchor token. 
The details are as follows:

\begin{itemize}[leftmargin=*,topsep=0.1em,itemsep=0.1em,parsep=0.1em]

\item \textbf{AnLLM-EP.} This approach uses punctuation marks within the sequence as anchor tokens. Punctuation marks, such as commas, periods, and question marks, are viewed as semantic boundaries within a sequence. As such, they can serve as anchor tokens in AnLLM. In our experiments of AnLLM-EP, we use the endpoint in English as the anchor tokens.

\item \textbf{AnLLM-AC.} This strategy entails the introduction of a new token to act as the sequence anchor. In our implementation, we designate <AC> as the new token and initialize its embedding using the mean value of the embedding matrix. 
For training data, we use the sentence tokenizer from the NLTK package to split texts into sentences, appending <AC> at the end of each sentence as the anchor token.\footnote{\url{https://www.nltk.org/api/nltk.tokenize.punkt.html}} 
During inference, <AC> tokens can be flexibly added to the text based on user requirements, such as adding one anchor for each demonstration, allowing for flexible and controllable sequence compression.


\end{itemize}


\subsection{Data and Training Procedure}


Considering that AnLLMs are expected to predict subsequent tokens within the context of keys/values hidden states of anchor tokens, this presents a significant challenge for existing open-source LLMs.
To this end, by substituting the self-attention networks with anchor-based self-attention networks as detailed in Section~\ref{sec:AbSAN}, we continually pre-train the Llama2 model using a publicly available corpus.



\paragraph{Data.} 
We employ the RedPajama-Data-1T-Sample dataset \cite{together2023redpajama} for the continuous pre-training  purpose.\footnote{\url{https://huggingface.co/datasets/togethercomputer/RedPajama-Data-1T-Sample}}
This dataset comprises $850,000$ samples with approximately $1$ billion tokens, which have been subjected to right truncation to fit the model context of $4,096$.

\paragraph{Training Procedure.} We train each model via the next token prediction objective on the dataset for one epoch, with a batch size of $512$.
The learning rate is set to $0.00002$ and constant after a linear warmup with $20$ update steps.
The AdamW \cite{loshchilov2018decoupled} with $\beta_{1}=0.9$ and $\beta_{2}=0.95$ is adopted as the gradient backtrack propagation optimizer.
All the training procedures are conducted with four $8 \times$ A100 GPU machines with $40$GB GPU Memory.

\paragraph{Training Loss and Perplexity.} 


The left-hand side of Figure~\ref{fig:trainnglossandppl} depicts the training loss associated with our models. The loss curves for AnLLM-EP and AnLLM-AC consistently decline to approximately 1.9, with AnLLM-AC achieving a lower loss. This observation suggests that continually pre-training an LLM using anchor-based attention masks is indeed viable, enabling the LLM to effectively learn the process of compressing sequence information into anchor tokens.

The right-hand side of Figure~\ref{fig:trainnglossandppl} displays the perplexity evaluation of the models with varying context lengths. Full attention is utilized to assess the language modeling capabilities of all models. Following the settings of \newcite{longlora}, the perplexity is evaluated on the test samples of the Proof-Pile datasets \cite{Rae2020Compressive}. 
The results demonstrate that both AnLLM-EP and AnLLM-AC models maintain a promising performance, exhibiting language modeling capacity comparable to the base model, Llama2-7B.
Moreover, this finding suggests that AnLLMs are compatible with full attention, as indicated by minimal perplexity decline.



\subsection{Evaluation}

In our investigation, we employ a diverse collection of benchmarks with varying text lengths to evaluate our outcomes, including OpenBookQA (OBQA) \cite{mihaylov2018can}, WinoGrande (WG) \cite{sakaguchi2021winogrande}, ARC-easy (ARC-e) and ARC-challenge (ARC-c) \cite{clark2018think}, PIQA \cite{bisk2020piqa}, HellaSwag (HS) \cite{zellers2019hellaswag}, SCIQ \cite{welbl-etal-2017-crowdsourcing}, and BoolQ \cite{clark-etal-2019-boolq}.
These benchmarks provide a comprehensive evaluation of various aspects, including reasoning, comprehension, understanding of the physical world, and predicting future events. Importantly, they cover texts of varying lengths, facilitating a thorough assessment of our model's performance across diverse tasks and text complexities, ranging from shorter input contexts in OBQA to longer texts in BoolQ. 
To measure the precision and efficiency of our models, we evaluate them across three dimensions using three distinct metrics for both zero-shot and five-shot settings. For AnLLM-AC in the five-shot setting, we incorporate the anchor token <AC> at the end of each demonstration.


\begin{itemize}[leftmargin=*,topsep=0.1em,itemsep=0.1em,parsep=0.1em]
\item \textbf{Accuracy (\emph{Acc}).} This conventional metric is utilized to gauge the prediction accuracy of models. In accordance with previous studies \cite{eval-harness}, we choose the options with the highest probabilities as predictions and calculate accuracy using the gold-standard labels.

\item \textbf{Keys/Values Caches Reduction (\textcolor{forestgreen}{$\mathcal{C}_{\Downarrow}$}).} 
In the context of the five-shot evaluation, the demonstrations can be cached in GPU memory for subsequent reuse. Nevertheless, extended demonstrations may require increased memory consumption. This metric is designed to assess the memory efficiency of the AnSAN technique.

\item \textbf{Inference Acceleration Ratio (\textcolor{blue}{$\mathcal{T}_{\Uparrow}$}). }
Similar to \newcite{wang-etal-2023-label}, capitalizing on the cached keys/values, we present the inference acceleration ratio, which serves as an indicator of the inference efficiency of the AnSAN technique.

\end{itemize}

Note that we first report full attention inference results for all models, then present results with the AnSAN method (+AnSAN) applied, compressing sequence information into anchor tokens.

\begin{table*}[t]

\centering
\begin{subtable}[h]{1.0\textwidth}
\centering
\begin{tabular}{lccccccccc}
\toprule
 &  \bf OBQA & \bf WG & \bf ARC-e & \bf ARC-c & \bf PIQA & \bf HS & \bf SCIQ & \bf BoolQ & \bf AVG.  \\ \midrule

\bf Llama2-7B & 31.4 & 69.1 & 76.3 & 43.4 & 78.1 & 57.1 & 93.7 & 77.7 & 65.8 \\ 

\bf AnLLM-EP & 33.2 & 68.0 & 73.4 & 40.8 & 77.8 & 55.0 & 94.4 & 74.4 & 64.6  \\ 

\bf AnLLM-AC & 31.6 & 68.5 & 74.4 & 42.5 & 78.3 & 54.7 & 93.8 & 77.0 & 65.1 \\ 
    
\bottomrule
\end{tabular}
\caption{The Zero-Shot Performance.}
\label{tab:0-shot}

\end{subtable}

\begin{subtable}[h]{1.0\textwidth}
\centering
\resizebox{\textwidth}{!}{
\begin{tabular}{lcccccccccc}
\toprule
 & & \bf OBQA & \bf WG & \bf ARC-e & \bf ARC-c & \bf PIQA & \bf HS & \bf SCIQ & \bf BoolQ & \bf AVG.  \\ \midrule

 & $L_{d}$ & 89 & 133 & 145 & 209 & 262 & 426 & 603 & 804 & 334 \\ 
  & $L_{x}$ & 18 & 26 & 36 & 42 & 42 & 90 & 130 & 169 & 69 \\  \midrule

\bf Llama2-7B & \emph{Acc} & 37.2 & 73.7 & 79.8 & 50.0 & 78.7 & 58.3 & 96.8 & 78.4 & 69.1 \\  \cdashline{1-11}[2.5pt/5pt]\noalign{\vskip 0.5ex} 
 \quad +AnSAN &  \emph{Acc} & 34.6 & 68.6 & 62.6 & 35.8 & 68.3 & 30.8 & 65.7 & 50.8 & 52.1 \\   \midrule

\bf AnLLM-EP & \emph{Acc} & 36.8 & 71.0 & 79.4 & 49.4 & 78.1 & 55.3 & 96.6 & 75.6 & 67.8 \\  \cdashline{1-11}[2.5pt/5pt]\noalign{\vskip 0.5ex} 
 \quad +AnSAN & \emph{Acc} & 36.2 & 68.0 & 76.7 & 45.6 & 78.2 & 52.6 & 93.1 & 74.0 & 65.6 \\ 
& $L_{kv}$ & 89 & 8 & 5 & 30 & 9 & 25 & 50 & 43 & 32 \\ 
& \textcolor{forestgreen}{$\mathcal{C}_{\Downarrow}$} & {\small $-$}0\% & \cellcolor{forestgreen!37}{\small $-$}94\% & \cellcolor{forestgreen!49}{\small $-$}97\% & \cellcolor{forestgreen!5}{\small $-$}86\% & \cellcolor{forestgreen!49}{\small $-$}97\% & \cellcolor{forestgreen!37}{\small $-$}94\% & \cellcolor{forestgreen!29}{\small $-$}92\% & \cellcolor{forestgreen!41}{\small $-$}95\% & \cellcolor{forestgreen!21}{\small $-$}90\% \\ 

& \textcolor{blue}{$\mathcal{T}_{\Uparrow}$} & {\small $\times$}1.0 & {\small $\times$}1.0 & {\small $\times$}1.0 & \cellcolor{blue!4}{\small $\times$}1.2 & \cellcolor{blue!8}{\small $\times$}1.4 & \cellcolor{blue!22}{\small $\times$}2.1 & \cellcolor{blue!32}{\small $\times$}2.6 & \cellcolor{blue!50}{\small $\times$}3.5 & \cellcolor{blue!14}{\small $\times$}1.7  \\

\midrule

\bf AnLLM-AC & \emph{Acc} & 37.2 & 72.3 & 79.8 & 49.0 & 78.6 & 56.9 & 96.8 & 77.5 & 68.5 \\  \cdashline{1-11}[2.5pt/5pt]\noalign{\vskip 0.5ex} 
 \quad +AnSAN &   \emph{Acc} & 35.6 & 70.6 & 79.2 & 47.9 & 78.7 & 55.6 & 95.7 & 76.6 & 67.5 \\ 
 & $L_{kv}$ & 5 & 5 & 5 & 5 & 5 & 5 & 5 & 5 & 5 \\ 
  & \textcolor{forestgreen}{$\mathcal{C}_{\Downarrow}$} & \cellcolor{forestgreen!37} {\small $-$}94\% & \cellcolor{forestgreen!48}{\small $-$}96\% & \cellcolor{forestgreen!49}{\small $-$}97\% & \cellcolor{forestgreen!50}{\small $-$}98\% & \cellcolor{forestgreen!50}{\small $-$}98\% & \cellcolor{forestgreen!50}{\small $-$}99\% & \cellcolor{forestgreen!50}{\small $-$}99\% & \cellcolor{forestgreen!50}{\small $-$}99\% & \cellcolor{forestgreen!50}{\small $-$}99\% \\

& \textcolor{blue}{$\mathcal{T}_{\Uparrow}$} & {\small $\times$}1.0 & {\small $\times$}1.0 & \cellcolor{blue!2}{\small $\times$}1.1 & \cellcolor{blue!4}{\small $\times$}1.2 & \cellcolor{blue!10}{\small $\times$}1.5 & \cellcolor{blue!20}{\small $\times$}2.0 & \cellcolor{blue!32}{\small $\times$}2.6 & \cellcolor{blue!50}{\small $\times$}3.5 & \cellcolor{blue!14}{\small $\times$}1.7  \\
    
\bottomrule
\end{tabular}
}
\caption{The Five-Shot Performance.}
\label{tab:5-shot}

\end{subtable}

\caption{Accuracy and Efficiency of LLMs on Question Answering Benchmarks.
\textcolor{forestgreen}{$\mathcal{C}_{\Downarrow}$} represents the reduction in keys/values cache size, while \textcolor{blue}{$\mathcal{T}_{\Uparrow}$} denotes the inference acceleration ratio during testing. \emph{Acc} stands for Accuracy. $L_{kv}$ represents the length of the keys/values cache. $L_{d}$ and $L_{x}$ denote the lengths of in-context learning demonstrations and input queries, respectively. Our methods effectively reduce cache sizes and boost inference efficiency.} 
\label{tab:mainresults}
\end{table*}

\section{Experimental Results}
\label{sec:results}

As evident from the results presented in Table~\ref{tab:mainresults}, both the AnLLM-AC and AnLLM-EP models demonstrate promising accuracy, comparable to that of the base model, while simultaneously improving memory and inference efficiency.

\paragraph{Accuracy (\emph{Acc}).} The proposed AnLLM-EP and AnLLM-AC models exhibit commendable accuracy across various benchmarks.

In the zero-shot setting, with full attention, AnLLM-EP and AnLLM-AC achieve average accuracies of 64.6\% and 65.1\%, respectively, comparable to Llama2-7B's 65.8\% accuracy. 
This suggests that training with integrated anchor tokens barely affects the model capacity, emphasizing the robustness of LLMs. Furthermore, our models excel in OBQA, PIQA, and SCIQ tasks.

In the five-shot setting, with five prior examples, AnLLM-EP and AnLLM-AC maintain dependable performance using full attention. 
When implementing the AnSAN technique, a slight accuracy decline across all models is observed. 
This is expected, as AnSAN, designed for memory efficiency, necessitates token removal, potentially leading to information loss. 
The degradation in BoolQ is most pronounced, which contains the longest demonstration tasks, indicating that the longer the text, the greater the information loss after compression. 
However, the average accuracy reduction is minimal, approximately 1.5\%, suggesting that AnSAN effectively balances memory-saving and model performance.

\paragraph{Keys/Values Cache Reduction (\textcolor{forestgreen}{$\mathcal{C}_{\Downarrow}$}).} The size of the keys/values cache is a critical factor in the practical implementation of LLMs, particularly concerning memory efficiency and computational resources. In this respect, the AnLLM-EP and AnLLM-AC models offer significant advantages. 

By adopting the AnSAN, these models are designed to dramatically reduce the keys/values cache size during inference. As shown in Table~\ref{tab:mainresults}, these models achieve remarkable reductions in cache size. Specifically, the average reduction percentages are around 90\% for AnLLM-EP and an impressive 99\% for AnLLM-AC. This is a substantial improvement compared to conventional approaches, which typically necessitate large cache sizes to store keys/values. 
These reductions in cache size translate to considerable savings in memory and computational resources, rendering these models highly efficient for practical applications.



\paragraph{Inference Acceleration Ratio (\textcolor{blue}{$\mathcal{T}_{\Uparrow}$}).} The inference acceleration ratio serves as a crucial metric reflecting the model's efficiency during the testing phase. By incorporating anchor tokens into natural language texts, we can repurpose the hidden states of anchor tokens as keys/values caches in the demonstrations, and then adopt an inference strategy as suggested by \newcite{wang-etal-2023-label}. In this scenario, both the AnLLM-EP and AnLLM-AC models demonstrate significant improvements.

Specifically, in the five-shot testing, both AnLLM-EP and AnLLM-AC models attain an average inference acceleration ratio of approximately 1.7 times. 
This represents a considerable advancement over the conventional non-caching method, which typically necessitates prolonged processing times due to the large number of tokens involved. 
As $L_{d}$ increases, reaching up to 3.5 times in the BoolQ task, the acceleration ratios also escalate, corroborating the findings of \newcite{wang-etal-2023-label}. 
This enhancement in processing speed leads to increased efficiency, making these models particularly apt for scenarios with limited resources.

\paragraph{The AnLLM-EP and AnLLM-AC models exhibit remarkable performance in natural language understanding benchmarks, effectively balancing accuracy, memory efficiency, and time acceleration.}
The incorporation of anchor tokens into AnLLMs, along with the utilization of the AnSAN technique for reducing keys/values cache size, allows these models to maintain performance on par while significantly improving memory efficiency and inference speed. 
The equilibrium achieved between model performance and computational efficiency is noteworthy and opens up new possibilities for the advancement of LLMs.

\section{Analysis}

To further elucidate our method's insights, we conduct a natural language generation experiment with the German-to-English (De2En) translation task. 
We evaluate the models using COMET-DA \cite{rei-etal-2022-comet}, indicating translation quality, and the Keys/Values Cache Reduction \textcolor{forestgreen}{$\mathcal{C}_{\Downarrow}$} metric, denoting memory efficiency as previously described. 
In line with previous findings, AnLLMs accept a minor accuracy trade-off (about 3 COMET-DA points) for enhanced memory efficiency. 
All LLMs are fine-tuned on the Alpaca dataset, combined with the newstest2017-2020 datasets, following \newcite{jiao2023parrot}. Results are presented in Table~\ref{tab:translation}.

\subsection{Compatibility and Flexibility of Full Attention and Anchor-based Attention}


The results offer significant insight into the interplay between anchor-based attention and full attention mechanisms in the De2En translation task. Since source sentences are vital in translation tasks, applying full attention to them is crucial for maintaining model performance. Thus, retaining the source sentence keys/values caches is expected to enhance AnLLM performance when implementing the AnSAN technique.
Specifically, when combining full attention with the AnSAN method, both AnLLM-EP and AnLLM-AC achieve approximately 80.0 COMET-DAE scores, comparable to other models using full attention exclusively. This indicates that the AnSAN technique is compatible with the full attention mechanism.
Consequently, our proposed models allow users to choose between full attention and anchor-based attention for input texts based on their needs, emphasizing the compatibility and flexibility of our models.

\subsection{Effective Cache Reduction for Real-Time Inference with the AnSAN Technique}
\label{sec:realtimeinfer}


The results in Table~\ref{tab:translation} show that our reduction strategy effectively minimizes keys/values caches during real-time inference. Specifically, as indicated in Line 15 of Algorithm~\ref{algo:anchorbasedinfer}, when generating an anchor token (i.e., the endpoint or <AC> tokens), our AnSAN-equipped models execute the reduction function to minimize the current keys/values caches. 
When discarding source sentence caches, we achieve approximately 77\% and 84\% reduction for the AnLLM-EP and AnLLM-AC models, respectively, albeit with a low COMET-DA score. However, when retaining source sentence caches, we still reduce around 44\% of caches for both models, achieving a COMET-DA score of approximately 80.0.
These results confirm the effectiveness of our anchor-based inference strategy for practical real-time inference applications.



\begin{table}[t!]
\centering
\resizebox{0.48\textwidth}{!}{
\begin{tabular}{lcccc}
\toprule
\bf Model & \bf Src Cache & \bf De2En & \bf MaxKV & \textcolor{forestgreen}{$\mathcal{C}_{\Downarrow}$} \\
\midrule
\bf Llama2-7b & \cmark & 83.1 & 220 & 0\% \\ \midrule
\bf AnLLM-EP & \cmark & 81.6 & 220 & 0\% \\  \cdashline{1-5}[2.5pt/5pt]\noalign{\vskip 0.5ex} 
\multicolumn{1}{l}{\multirow{2}{*}{\quad +AnSAN}} & \xmark & 78.5 & 50 & 77\% \\
& \cmark & 80.3 & 124 & 44\% \\ \midrule
\bf AnLLM-AC & \cmark & 82.4 & 220 & 0\% \\ \cdashline{1-5}[2.5pt/5pt]\noalign{\vskip 0.5ex} 
\multicolumn{1}{l}{\multirow{2}{*}{\quad +AnSAN}} & \xmark & 78.0 & 35 & 84\%  \\
& \cmark & 80.0 & 125 & 43\%  \\ 
\bottomrule
\end{tabular}
}
\caption{COMET-DA Scores and Keys/Values Cahces for the WMT23 German-to-English (De2En) Translation Task. The term ``Src Cach'' denotes retaining source sentence hidden states in Keys/Values Caches, while ``MaxKV'' refers to the average maximum keys/values length during inference.}
\label{tab:translation}
\end{table}





\begin{table*}[ht]
\resizebox{\textwidth}{!}{
\begin{tabular}{lccccccccc}
\toprule
\bf Settings              & \bf OBQA & \bf WG   & \bf ARC-e & \bf ARC-c & \bf PIQA & \bf HS   & \bf SCIQ & \bf BoolQ & \bf AVG. \\ \midrule
every-10-tokens     & 21.4 & 69.2 & 63.6  & 33.3 & 75.0 & 48.1 & 81.9 & 65.4  & 57.6 \\
random-prob-0.1       & 21.6 & 69.9 & 64.8  & 34.8  & 75.5 & 48.1 & 80.0   & 67.4  & 57.8 \\ 
every-demonstration & \bf 35.6 & \bf 70.6 & \bf 79.2  & \bf 47.9  & \bf 78.7 & \bf 55.6 & \bf 95.7 & \bf 76.6  & \bf 67.5 \\
\bottomrule
\end{tabular}}
\caption{Accuracy on Question Answering Benchmarks with Different Anchor Positions. All the experiments are conducted with the AnLLM-AC model in Table~\ref{tab:mainresults}.}
\label{tab:ancorpos}
\end{table*}

\section{Ablation Study}

\subsection{Impact of Anchor Positions}

An intriguing question arises regarding the impact of anchor positions on model performance. 
In this section, we investigate the effects of varying anchor positions using the AnLLM-AC model, which enables us to modify the anchor position.
Specifically, we employ the data settings from Section~\ref{sec:AbSAN} and examine three position settings: the first compresses every 10 tokens, the second applies random compression, and the third compresses each demonstration, consistent with the setting in Table~\ref{tab:mainresults}. 
For the second setting, an anchor token is randomly inserted after each token with a probability of 0.1. The experimental results are presented in Table~\ref{tab:ancorpos}.

Accordingly, we observe that the choice of anchor positions significantly affects the model's performance across various question-answering benchmarks. The "every-demonstration" setting consistently outperforms the other two settings, achieving the highest average accuracy of 67.5\%. This suggests that strategically placing anchors at semantically meaningful positions, such as after each demonstration, can effectively enhance the model's ability to capture and utilize the information contained in the input texts.

In comparison, the ``every-10-tokens'' and ``random-prob-0.1'' settings yield lower average accuracies of 57.6\% and 57.8\%, respectively. These results indicate that compressing input texts at fixed intervals or randomly inserting anchor tokens may not be as effective in facilitating the model's understanding and reasoning processes. The suboptimal performance of these settings could be attributed to the potential loss of semantic coherence and structural information as a result of arbitrary compression or random anchor placement.

Overall, our ablation study highlights the importance of carefully selecting anchor positions in the AnLLM-AC model to maximize its performance on question-answering tasks. The superior performance of the "every-demonstration" setting demonstrates the benefit of aligning anchor positions with semantically meaningful boundaries in the input texts. Future research could explore more sophisticated strategies for anchor placement, taking into account the linguistic and contextual properties of the input data to further improve the model's performance on complex reasoning tasks.

\subsection{Training from Scratch}

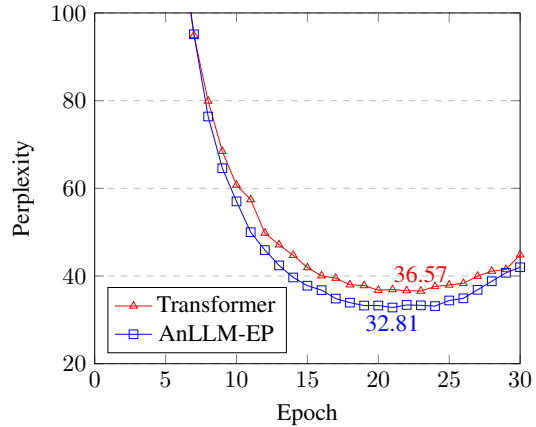
\begin{figure}
    \centering
\resizebox{0.45\textwidth}{!}{

\begin{tikzpicture}
\begin{axis}[
    xlabel={Epoch},
    ylabel={Perplexity},
    xmin=0, xmax=30,
    ymin=20, ymax=100,
    xtick={0,5,10,15,20,25,30},
    ytick={0,20,40,60,80,100,120},
    legend pos=south west,
    ymajorgrids=true,
    grid style=dashed,
    nodes near coords,
    point meta=explicit symbolic
]

\addplot[
    color=red,
    mark=triangle,
    nodes near coords style={yshift=0pt},
    mark options={scale=1}
] coordinates {
    (1,1107.19)(2,485.29)(3,299.85)(4,208.66)(5,152.53)(6,117.07)(7,94.92)(8,79.92)(9,68.45)(10,60.74)(11,57.43)(12,49.80)(13,47.12)(14,44.73)(15,41.92)(16,40.06)(17,39.51)(18,38.00)(19,37.82)(20,36.76)(21,36.91)(22,36.66)(23,36.57)[36.57](24,37.61)(25,37.95)(26,38.34)(27,39.97)(28,41.12)(29,41.49)(30,44.83)
};
\addlegendentry{Transformer}

\addplot[
    color=blue,
    mark=square,
    nodes near coords style={yshift=-15pt},
    mark options={scale=1}
] coordinates {
    (1,1107.70)(2,557.04)(3,326.66)(4,226.83)(5,159.87)(6,118.25)(7,95.14)(8,76.38)(9,64.59)(10,57.02)(11,50.00)(12,45.92)(13,42.42)(14,39.67)(15,37.79)(16,36.78)(17,34.85)(18,33.92)(19,33.26)(20,33.27)(21,32.81)[32.81](22,33.40)(23,33.31)(24,33.14)(25,34.42)(26,34.91)(27,36.85)(28,38.80)(29,40.73)(30,42.00)
};
\addlegendentry{AnLLM-EP}

\end{axis}
\end{tikzpicture}
}

    \caption{Perplexity Comparison Across Epochs for Small Standard Transformer and AnLLM-EP Models.}
    \label{fig:tscratch}
\end{figure}


To evaluate the language modeling capabilities of our anchor-based language model (AnLLM-EP), we perform a comparison with the standard Transformer model. This comparison involves the training of compact models from scratch, using the Wikitext-103 dataset.\footnote{\url{https://huggingface.co/datasets/iohadrubin/wikitext-103-raw-v1}} 
Each model is configured with 18 layers, 4096 hidden states, and 16 heads. 
As shown in Figure~\ref{fig:tscratch}, the AnLLM-EP model notably outperforms the standard Transformer model, achieving a lower perplexity of 32.81, compared to 36.57.
This notable outcome suggests that the anchor-based training approach may enhance the effectiveness of language modeling tasks. 
In future research, it would be intriguing to further investigate the potential advantages of the anchor-based training strategy for training LLMs from scratch.

\section{Conclusion}

LLMs have emerged as a significant research area in the field of artificial intelligence. 
However, despite their exceptional performance across various natural language tasks, the practical application of these models is limited by their significant memory overhead and time efficiency. 
Implementing LLMs on resource-constrained devices, such as smartphones, poses a unique challenge.
To address this issue, we propose anchor-based LLMs with the AnSAN technique. Our experiments demonstrate that by sacrificing a marginal 1.5\% in precision, our approach saves 99\% of keys/values cache memory while simultaneously improving inference speed by up to 3.5 times. 
Our methods' application in machine translation showcases their compatibility and flexibility, effectively enhancing memory efficiency for practical use.
Our novel approach is practical, straightforward, flexible, and compatible with existing methods, paving the way for further adoption of LLMs in real-world applications.


\section*{Limitations}

While our proposed AnLLMs demonstrate significant improvements in memory efficiency and inference acceleration, there are several limitations to consider:

\begin{enumerate}
    \item \textbf{Accuracy Trade-off:} As observed in the experimental results, our method incurs a minor decrease in accuracy (within 1.5\%) compared to the original model. This limitation stems from the information compression process, which may lead to information loss. Despite its minimal impact, this trade-off should be considered in practical applications. In future works, additional evaluation methods could further enrich our assessment \cite{ye2024benchmarking,wang2023findings}.
        
    \item \textbf{Applicability to Various Tasks:} Our experiments primarily focus on question-answering benchmarks and machine translation tasks. The effectiveness of our method in other NLP tasks and domains remains to be thoroughly investigated. Future work will explore the applicability and performance of our method across a broader range of tasks and domains \cite{zhao-etal-2023-adaptive,wang2023document,pang-etal-2024-monmt-modularly,wang2024benchmarking}.

    
    \item \textbf{Optimal Anchor Token Selection:} In our implementation, we chose the last token in a sequence as the anchor token. However, the optimal anchor token selection may vary across different tasks and domains. Further research is needed to develop more sophisticated strategies for identifying and leveraging the most suitable anchor tokens.
    
    \item \textbf{Scalability to Other LLMs:} We have applied our method to the open-source Llama2 models. It remains to be seen how our approach would perform when applied to other open-source LLMs, such as Falcon and Qwen \cite{almazrouei2023falcon,bai2023qwen}. Evaluating the effectiveness and scalability of our method on more extensive language models is an essential direction for future research.
    
\end{enumerate}

Despite these limitations, our work presents a novel approach to enhancing memory efficiency and inference acceleration in LLMs. Future research efforts should address these limitations, refining our method and extending its applicability to a wider range of tasks and model architectures.

\section*{Ethics Statement}

We place great importance on ethical considerations and rigorously adhere to the ACL Ethics Policy. 
In this paper, we propose an anchor-based large language model that reduces the Keys/Values cache size and enhances inference speed during the inference stage. 
The resources and methods employed in this paper are publicly accessible and have been extensively adopted by researchers in the field of large language models. We ensure that the findings and conclusions presented in this paper are reported accurately and objectively.

\section*{Acknowledgments}

This work was supported in part by the Science and Technology Development Fund, Macau SAR (Grant Nos. FDCT/0070/2022/AMJ, FDCT/060/2022/AFJ), Ministry of Science and Technology of China (Grant No. 2022YFE0204900), National Natural Science Foundation of China (Grant No. 62261160648), the Multi-year Research Grant from the University of Macau (Grant No. MYRG-GRG2023-00006-FST-UMDF), and Tencent AI Lab Rhino-Bird Gift Fund (Grant No. EF2023-00151-FST). This work was performed in part at SICC which is supported by SKL-IOTSC, and HPCC supported by ICTO of the University of Macau.

\bibliography{custom,anthology}

\clearpage

\appendix



\section{More Experimental Results}

\subsection{Detailed Perpelixity Evaluation}

In this section, we present a comprehensive analysis of the perplexity evaluation results, as illustrated in Table~\ref{tab:anchorppl}. The perplexity scores were calculated for various evaluation context lengths, ranging from 256 to 4096 tokens, utilizing the Proof-Pile datasets \cite{Rae2020Compressive}. The table compares the performance of Llama2-7B, AnLLM-AC, AnLLM-EP, and their corresponding variants incorporating the AnSAN technique.
Our findings reveal that the AnSAN technique, on average, leads to a one-point increase in perplexity, which negatively impacts the modeling capabilities of the models to some extent. These outcomes resonate with the trade-off between accuracy and efficiency observed in Table~\ref{tab:mainresults}.

\subsection{Testing Acceleration Ratio to Full-Caching Method}

In Section~\ref{sec:results}, we report the testing acceleration ratio following the setting of \newcite{wang-etal-2023-label}, comparing the time difference between caching and non-caching inference. 
Although our method reduces the keys/values caches, enabling smaller space for prefix information storage and improving testing time up to {\small{$\times$}}3.5, we are still curious about whether it would enhance inference efficiency if conventional methods use full caches that save all keys/values of prefix tokens. 
As a supplement to Table~\ref{tab:mainresults}, we present the testing acceleration ratio between anchor-caching and full-caching inference in Table~\ref{tab:testingtimecaching}.
The acceleration ratios for AnLLM-EP-AnSAN and AnLLM-AC-AnSAN achieve the highest improvements observed in tasks such as HS, SCIQ, and BoolQ. The average acceleration ratios for AnLLM-EP-AnSAN and AnLLM-AC-AnSAN are 1.03.
These results demonstrate that our anchor-based caching method can enhance inference efficiency even when compared to conventional methods that save all keys/values of prefix tokens.
These results suggest that our anchor-based caching approach, which saves only the keys/values caches of anchor tokens, can effectively accelerate the inference process for the lengthy prefix texts.

\subsection{Model Scalability Assessment}

To examine the scalability of our approach, we extend the AnLLM-AC model to 13B and assess its performance on eight question-answering benchmarks using the same evaluation strategy as previously mentioned. 
In comparison to the 7B AnLLM models in Table~\ref{tab:mainresults}, Results in Table~\ref{tab:13bmmlu} indicate that as the model size expands, the AnLLM-AC model achieves accuracies of 67.5\% and 70.0\% for 0-shot and 5-shot testing, respectively, resulting in up to a 2.4\% improvement. 
Moreover, by incorporating anchor-based attention, the AnLLM-AC-AnSAN model achieves an average accuracy of 69.5\%, signifying a 2.0\% increase. The performance enhancement underscores the effectiveness of our methods in accommodating larger model capacities. 
The consistent improvements observed in the AnLLM-AC model across various scenarios highlight its robustness and adaptability. Furthermore, the increased performance of the AnLLM-AC-AnSAN model, facilitated by anchor-based attention, emphasizes the potential of our approaches in optimizing LLMs. 
Collectively, these findings point to promising avenues for future research aimed at maximizing the utility and efficiency of AnLLM.

\subsection{Case Study in Real-Time Inference}

To elaborate on the optimization of keys/values caches by AnLLM-EP and AnLLM-AC during real-time inference, we reference examples from the translation task in Section~\ref{sec:realtimeinfer}. As per Table~\ref{tab:translationexample}, AnLLM-EP and AnLLM-AC use "endpoints" (".") and "<AC>" tokens as anchor tokens, respectively.
During inference, both models employ auto-regressive generation, creating outputs token-by-token. Upon generating an anchor token (as per Line 16, Algorithm~\ref{algo:anchorbasedinfer}), the Reduction function (defined in Line 1) is activated, preserving relevant caches and eliminating others.
As a result, the Keys/Values Cache lengths are reduced to roughly the sequence length, averaging around 50 for AnLLM-EP and 35 for AnLLM-AC, as shown in Table~\ref{tab:translation}.

\subsection{Attention Pattern in AnLLMs}

\begin{figure*}[ht]
    \centering
    \includegraphics{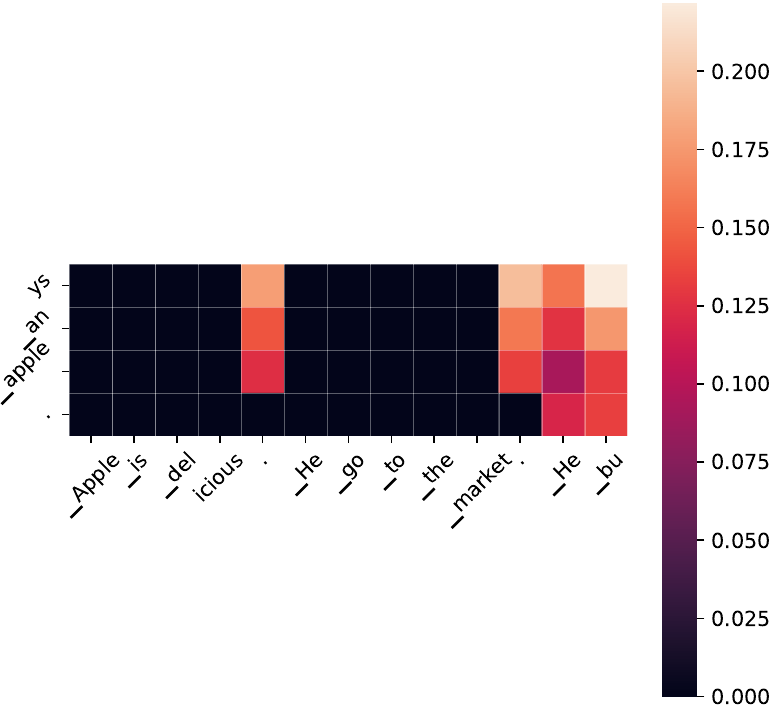}
    \caption{Case Study on Attention Pattern. With the AnLLM-EP model, the heatmap presents the average correlation between tokens across layers of output hidden states.}
    \label{fig:attp}
\end{figure*}

Regarding the attention pattern, we have conducted a case study using the AnLLM-EP model. 
As shown in Figure~\ref{fig:attp}, given the sentence ``Apple is delicious. He goes to the market. He buys an apple.'', 
we detokenize and split it into two segments: ``\_Apple \_is \_del icious . \_He \_go \_to \_the \_market . \_He \_b'' and ``ys \_an \_apple .''.
By employing a heatmap to visualize the attention pattern between the latter segment and the former, we observe that the token ``ys'' attends more to the second endpoint, which compresses the information of ``He goes to the market.''. This is a reasonable and interesting finding, as ``ys'' is part of the word ``buys''. Additionally, the token ``\_apple'' attends more to the first endpoint, which compresses the information of ``Apple is delicious.''. These attention patterns offer some interpretability for our method.

\section{Data Settings}
\label{sec:appendixa}

To provide a thorough insight into how we continually pre-train the model into AnLLM and carry out evaluations, we showcase some data examples in this section for both training and testing data.

\subsection{Training Data Examples}


In this section, we provide examples to illustrate the specific data format used in training the AnLLM models. For the AnLLM-EP model, the endpoints act as anchor tokens, allowing us to directly utilize natural language texts. For the AnLLM-AC model, we append a new token <AC> at the end of each sequence in the input texts, which are initially split into sentences using the NLTK toolkits.\footnote{\url{https://www.nltk.org/api/nltk.tokenize.punkt.html}} Some examples are presented in Table~\ref{tab:trainingdata}.
All the trainig data are downloaded from HuggingFace\footnote{\url{https://huggingface.co/datasets/togethercomputer/RedPajama-Data-1T-Sample}}, an open-source community.

\subsection{Testing Data Examples}


For the testing outlined in the results section (Section~\ref{sec:results}), we employ the same evaluation method as in previous work \cite{eval-harness}, which treats each choice as text generation and computes the corresponding probabilities, respectively. Table~\ref{tab:testingdata} presents some evaluation examples.

\begin{table*}[ht]
    \centering
\begin{tabular}{lcccccc}
\toprule
     \multirow{2}{*}{\bf Method}   & \multicolumn{6}{c}{\bf Evaluation Context Length} \\ \cmidrule(r){2-7}
    &  \textbf{256} & \textbf{512} & \textbf{1024} & \textbf{2048} & \textbf{4096} & \textbf{AVG.}  \\ \midrule 
    \bf Llama2-7B & 5.42 & 4.32 & 3.64 & 3.2  & 2.91 & 3.90       \\ \midrule

    \bf AnLLM-AC &  5.70 & 4.53 & 3.81 & 3.36 & 3.07 & 4.09  \\ 
    \quad+AnSAN &   6.61 & 5.61 & 5.01 & 4.62 & 4.40 & 5.25  \\ \midrule

    \bf AnLLM-EP & 5.62 & 4.44 & 3.73 & 3.32 & 3.04 & 4.04  \\ 
    \quad+AnSAN & 6.18 & 5.17 & 4.62 & 4.31 & 4.14 & 4.88 \\ 
    
\bottomrule
\end{tabular}
    \caption{Perplixity Evaluation without or with the AnSAN Technique. The test samples are from the Proof-Pile dataset.}
    \label{tab:anchorppl}
\end{table*}

\begin{table*}[ht]
\centering
\resizebox{1.0\textwidth}{!}{
\begin{tabular}{lccccccccc}
\toprule
 &  \bf OBQA & \bf WG & \bf ARC-e & \bf ARC-c & \bf PIQA & \bf HS & \bf SCIQ & \bf BoolQ & \bf AVG.  \\ \midrule
AnLLM-EP-AnSAN & {\small $\times$}1.00 & {\small $\times$}1.00 & {\small $\times$}1.00  & {\small $\times$}1.00  & {\small $\times$}1.00 & {\small $\times$}1.06 & {\small $\times$}1.14 & {\small $\times$}1.13  & {\small $\times$}1.03 \\
 
AnLLM-AC-AnSAN  & {\small $\times$}1.00 & {\small $\times$}1.02 & {\small $\times$}1.00  & {\small $\times$}1.00  & {\small $\times$}1.00 & {\small $\times$}1.01 & {\small $\times$}1.10 & {\small $\times$}1.13  & {\small $\times$}1.03 \\
\bottomrule
\end{tabular}}
\caption{Testing Acceleration Ratio on Question-Answering Tasks between Anchor-Caching and Full-Caching Inference with Five-Shot Demonstrations. Anchor-caching refers to saving only the keys/values caches of anchor tokens with the AnSAN technique, while full-caching denotes saving caches for all prefix tokens. The tasks are arranged according to the demonstration lengths. The experiments are the same as those of Table~\ref{tab:mainresults}. These results suggest that inference speed differences for short texts are minimal but become more pronounced for longer texts. However, full-caching inference demands more GPU memory to store the complete keys/values caches, which is not ideal for environments with limited computational resources.}
\label{tab:testingtimecaching}
\end{table*}

\begin{table*}[ht]
\centering
\begin{tabular}{lccccccccc}
\toprule
 \bf Model &  \bf OBQA & \bf WG & \bf ARC-e & \bf ARC-c & \bf PIQA & \bf HS & \bf SCIQ & \bf BoolQ & \bf AVG.  \\ \midrule
\multicolumn{8}{l}{\textit{Zero-Shot Performance }} \\ \cdashline{1-10}[2.5pt/5pt]\noalign{\vskip 0.5ex} 

\bf Llama2-7B & 31.4 & 69.1 & 76.3 & 43.4 & 78.1 & 57.1 & 93.7 & 77.7 & 65.8 \\ 

\bf Llama2-13b             & 35.2 & 72.1 & 79.4  & 48.5  & 79.1 & 60.0 & 94.5 & 80.6  & 68.7 \\ 
\bf AnLLM-AC-7B & 31.6 & 68.5 & 74.4 & 42.5 & 78.3 & 54.7 & 93.8 & 77.0 & 65.1 \\ 
\bf AnLLM-AC-13B & 35.2 & 70.7 & 77.9  & 46.9  & 78.6 & 58.1 & 94.7 & 78.1  & 67.5 \\ \midrule
\multicolumn{8}{l}{\textit{Five-Shot Performance }} \\ \cdashline{1-10}[2.5pt/5pt]\noalign{\vskip 0.5ex} 
\bf Llama2-7B & 37.2 & 73.7 & 79.8 & 50.0 & 78.7 & 58.3 & 96.8 & 78.4 & 69.1 \\  
\bf Llama2-13b             & 38.2 & 76.3 & 82.2  & 52.6  & 80.0 & 61.4 & 97.5 & 83.5  & 71.5 \\

\bf AnLLM-AC-7B & 37.2 & 72.3 & 79.8 & 49.0 & 78.6 & 56.9 & 96.8 & 77.5 & 68.5 \\  
 \quad +AnSAN &  35.6 & 70.6 & 79.2 & 47.9 & 78.7 & 55.6 & 95.7 & 76.6 & 67.5 \\ 

\bf AnLLM-AC-13B               & 36.6 & 72.5 & 81.6  & 53.7  & 79.2 & 59.6 & 97.5 & 79.6  & 70.0 \\
\quad +AnSAN & 36.0 & 74.0 & 81.6  & 52.0  & 79.1 & 58.4 & 96.3 & 78.8  & 69.5 \\

\bottomrule
\end{tabular}
\caption{Accuracy of 13B LLMs on Question Answering Benchmarks. Compared to 7B AnLLMs, the 13B AnLLMs exhibit superior performance, with up to 2.0 accuracy enhancements, suggesting that AnLLMs possess excellent scalability to larger model architectures.}
\label{tab:13bmmlu}

\end{table*}







\begin{table*}[ht]
\centering
\begin{subtable}[h]{1.0\textwidth}
\centering
\resizebox{\textwidth}{!}{
    \begin{tabular}{cp{14cm}} 
    \toprule
          \bf Input   &  Below is an instruction that describes a task, paired with an input that provides further context. Write a response that appropriately completes the request. \\ \cdashline{2-2}\noalign{\vskip 0.5ex}
             & \#\#\# Instruction: Translate the following sentences from German to English. \\ \cdashline{2-2}\noalign{\vskip 0.5ex}
             & \#\#\# Input: Nachdem Werte in einen anderen Teil des Speichers eingeschrieben wurden, wird das CMOS RAM in der gleichen Weise wie das Communications RAM geprüft. Wurde der Test bestanden, werden alle Speicherstellen auf ihren früheren Wert eingestellt. LED-Wert: 00 0011 Wenn das Gerät mit dem o.a. Display hält, liegt ein Fehler vor. Prüfen Sie in diesem Falle U 85 und U 86 und die damit verbundenen Stromkreise bzw. die Dekodierung.  \\ \cdashline{2-2}\noalign{\vskip 0.5ex}
             & \#\#\# Response: \\ \cdashline{1-2}\noalign{\vskip 0.5ex}

        \bf Output & After values have been written to another part of the CMOS RAM, the CMOS RAM is tested in the same way as the communications RAM. If the test is successful, all storage locations will be reset to their former value. LED value: 00 0011 If the device is displaying this value, there is a fault to be found. In this case, check U85 and U86 and the associated power circuits, as well as decoding. \\ \midrule
        COMET-DA & 82.2 \\
        Length & 293 \\
        MaxKV & 170 \\
        \textcolor{forestgreen}{$\mathcal{C}_{\Downarrow}$} & 42\% \\
        \bottomrule
            
    \end{tabular}}
\caption{An Example of the AnLLM-EP Model in De2En Translation Task.}
\end{subtable}

\begin{subtable}[h]{1.0\textwidth}
\centering
\resizebox{\textwidth}{!}{

    \begin{tabular}{cp{14cm}} 
    \toprule
          \bf Input   &  Below is an instruction that describes a task, paired with an input that provides further context. Write a response that appropriately completes the request. <AC> \\ \cdashline{2-2}\noalign{\vskip 0.5ex}
             & \#\#\# Instruction: Translate the following sentences from German to English. <AC> \\ \cdashline{2-2}\noalign{\vskip 0.5ex}
             & \#\#\# Input: Der Spielablauf wird ausgelöst, wenn der Druckschalter START betätigt wird. <AC>  Der LED-Schalter bleibt erleuchtet und die Spielgeräusche werden hörbar. <AC>  Wird der letzte Spieler-Panzer vom feindlichen Angreifer eliminiert, ist ein Spiel beendet. <AC>  Das Spiel beginnt mit langsam fahrenden Panzern, die auf dem Bildschirm sichtbar werden. <AC>  Die Position der jeweiligen feindlichen Panzer oder Raketen kann der Spieler dadurch bestimmen, dass er den eingeblendeten Text im oberen linken Hinweisdisplay-Winkel seines Cockpits beachtet. <AC>  Ein Hinweis kann z.B. <AC>  lauten: GEGNER VON RECHTS. <AC>  Zusätzlich wird auf dem weitreichenden Radarschirm, begleitet von einem Sonar-BLIP-Ton, ein Punkt sichtbar, der die feindliche Stellung darstellt. <AC>  \\ \cdashline{2-2}\noalign{\vskip 0.5ex}
             & \#\#\# Response: \\ \cdashline{1-2}\noalign{\vskip 0.5ex}

        \bf Output & The game is triggered when the start switch is pressed. <AC>  The LED switch remains illuminated and the game sounds can be heard. <AC>  When the last enemy player's tank is destroyed, the game is over. <AC>  The game begins with tanks moving slowly move across the screen. <AC>  The player's tank's position can be determined by the player by paying attention to the respective enemy tank or missile icon in the upper left corner of his cockpit's display screen. <AC>  An example of an instruction can be: ENEMY ON THE RIGHT. <AC>  In addition, on the wide-range radar screen, accompanied by a sonar BLIP sound, a point is visible, which indicates the enemy position. \\ \midrule
        COMET-DA & 81.8 \\
        Length & 431 \\
        MaxKV & 272 \\
        \textcolor{forestgreen}{$\mathcal{C}_{\Downarrow}$} & 37\% \\
        \bottomrule
            
    \end{tabular}}
\caption{An Example of the AnLLM-AC Model in De2En Translatio Task.}
\end{subtable}

\caption{Case Study of Real-time Inference. During the inference process, AnLLM-EP generates "endpoint" as the anchor token, whereas AnLLM-AC produces "<AC>" as the anchor token. Once upon an anchor token, we execute the {R\small{EDUCTION}} as shown in Line 16 to reduce the keys/values caches.}
\label{tab:translationexample}

\end{table*}


\begin{table*}[!t]
\centering
\begin{subtable}[h]{1.0\textwidth}
\centering
\begin{tabular}{l}
\toprule
\fbox{
\begin{minipage}[t]{0.95\textwidth}%
Gender diversity, or more often its lack thereof, among participants to software development activities has been thoroughly studied in recent years\textcolor{red}{.} In particular, the presence of, effects of, and countermeasures for gender bias in Free/Open Source Software (FOSS) have received a lot of attention over the past decade\textcolor{red}{.} Geographic diversity is on the other hand the kind of diversity that stems from participants in some global activity coming from different world regions and cultures\textcolor{red}{.} Geographic diversity in FOSS has received relatively little attention in scholarly works\textcolor{red}{.} In particular, while seminal survey-based and point-in-time medium-scale studies of the geographic origins of FOSS contributors exist, large-scale longitudinal studies of the geographic origin of FOSS contributors are still lacking\textcolor{red}{.} Such a quantitative characterization would be useful to inform decisions related to global development teams and hiring strategies in the information technology (IT) market, as well as contribute factual information to the debates on the economic impact and sociology of FOSS around the world\textcolor{red}{.} ...
\end{minipage}}\tabularnewline
\bottomrule
\end{tabular}
\caption{A Training Data Example for the AnLLM-EP Model. The endpoints in the text serve as the anchor tokens.}
\end{subtable}

\begin{subtable}[h]{1.0\textwidth}
\centering
\begin{tabular}{l}
\toprule
\fbox{
\begin{minipage}[t]{0.95\textwidth}%
Gender diversity, or more often its lack thereof, among participants to software development activities has been thoroughly studied in recent years. \textcolor{red}{<AC>} In particular, the presence of, effects of, and countermeasures for gender bias in Free/Open Source Software (FOSS) have received a lot of attention over the past decade. \textcolor{red}{<AC>} Geographic diversity is on the other hand the kind of diversity that stems from participants in some global activity coming from different world regions and cultures. \textcolor{red}{<AC>} Geographic diversity in FOSS has received relatively little attention in scholarly works. \textcolor{red}{<AC>} In particular, while seminal survey-based and point-in-time medium-scale studies of the geographic origins of FOSS contributors exist, large-scale longitudinal studies of the geographic origin of FOSS contributors are still lacking. \textcolor{red}{<AC>} Such a quantitative characterization would be useful to inform decisions related to global development teams and hiring strategies in the information technology (IT) market, as well as contribute factual information to the debates on the economic impact and sociology of FOSS around the world. \textcolor{red}{<AC>} ...
\end{minipage}}\tabularnewline
\bottomrule
\end{tabular}
\caption{A Training Data Example for the AnLLM-AC Model. The newly added tokens <AC> in the text serve as the anchor tokens.}
\end{subtable}

\caption{Training Data Examples for the AnLLM-EP and AnLLM-AC models. For the AnLLM-EP model, the endpoints are the natural anchor tokens. For the AnLLM-AC model, we manually append <AC> tokens to sequences as the anchor tokens.}
\label{tab:trainingdata}

\end{table*}

\begin{table*}[!t]
\centering
\begin{subtable}[h]{1.0\textwidth}
\centering
\begin{tabular}{l}
\toprule
\fbox{
\begin{minipage}[t]{0.95\textwidth}%
\textbf{Choice 1:} Slacklining: A group of people have stretched a tightrope across a gym. They \textit{\textcolor{red}{take turns trying to balance and walk on the rope.}} \\
\textbf{Choice 2:} Slacklining: A group of people have stretched a tightrope across a gym. They \textit{\textcolor{red}{slide down with it, jumping and spinning in the air.}} \\
\textbf{Choice 3:} Slacklining: A group of people have stretched a tightrope across a gym. They \textit{\textcolor{red}{cross it together, swinging back and fourth in anticipation.}} \\
\textbf{Choice 4:} Slacklining: A group of people have stretched a tightrope across a gym. They \textit{\textcolor{red}{drop an orange rope at the end.}}
\end{minipage}}\tabularnewline
\bottomrule
\end{tabular}
\caption{A Zero-Shot Testing Data Example of the HellaSwag Task. The log-likelihood of the red texts is computed as the choice probabilities.}
\end{subtable}

\begin{subtable}[h]{1.0\textwidth}
\centering
\begin{tabular}{l}
\toprule
\fbox{
\begin{minipage}[t]{0.95\textwidth}%
\textbf{Choice 1:} Ballet: We see a pregnant lady doing ballet in a studio. The lady spins and does a pliea. Demonstration 2 Demonstration 3 Demonstration 4 Demonstration 5 Slacklining: A group of people have stretched a tightrope across a gym. They \textit{\textcolor{red}{take turns trying to balance and walk on the rope.}} \\
\textbf{Choice 2:} Ballet: We see a pregnant lady doing ballet in a studio. The lady spins and does a pliea. Demonstration 2 Demonstration 3 Demonstration 4 Demonstration 5 Slacklining: A group of people have stretched a tightrope across a gym. They \textit{\textcolor{red}{slide down with it, jumping and spinning in the air.}} \\
\textbf{Choice 3:} Ballet: We see a pregnant lady doing ballet in a studio. The lady spins and does a pliea. Demonstration 2 Demonstration 3 Demonstration 4 Demonstration 5 Slacklining: A group of people have stretched a tightrope across a gym. They \textit{\textcolor{red}{cross it together, swinging back and fourth in anticipation.}} \\
\textbf{Choice 4:} Ballet: We see a pregnant lady doing ballet in a studio. The lady spins and does a pliea. Demonstration 2 Demonstration 3 Demonstration 4 Demonstration 5 Slacklining: A group of people have stretched a tightrope across a gym. They \textit{\textcolor{red}{drop an orange rope at the end.}}
\end{minipage}}\tabularnewline
\bottomrule
\end{tabular}
\caption{A Five-Shot Testing Data Example of the HellaSwag Task for the ALLM-EP Inference.}
\end{subtable}

\begin{subtable}[h]{1.0\textwidth}
\centering
\begin{tabular}{l}
\toprule
\fbox{
\begin{minipage}[t]{0.95\textwidth}%
\textbf{Choice 1:} Ballet: We see a pregnant lady doing ballet in a studio. The lady spins and does a pliea. <AC> Demonstration 2 <AC> Demonstration 3 <AC> Demonstration 4 <AC> Demonstration 5 <AC> Slacklining: A group of people have stretched a tightrope across a gym. They \textit{\textcolor{red}{take turns trying to balance and walk on the rope.}} \\
\textbf{Choice 2:} Ballet: We see a pregnant lady doing ballet in a studio. The lady spins and does a pliea. <AC> Demonstration 2 <AC> Demonstration 3 <AC> Demonstration 4 <AC> Demonstration 5 <AC> Slacklining: A group of people have stretched a tightrope across a gym. They \textit{\textcolor{red}{slide down with it, jumping and spinning in the air.}} \\
\textbf{Choice 3:} Ballet: We see a pregnant lady doing ballet in a studio. The lady spins and does a pliea. <AC> Demonstration 2 <AC> Demonstration 3 <AC> Demonstration 4 <AC> Demonstration 5 <AC> Slacklining: A group of people have stretched a tightrope across a gym. They \textit{\textcolor{red}{cross it together, swinging back and fourth in anticipation.}} \\
\textbf{Choice 4:} Ballet: We see a pregnant lady doing ballet in a studio. The lady spins and does a pliea. <AC> Demonstration 2 <AC> Demonstration 3 <AC> Demonstration 4 <AC> Demonstration 5 <AC> Slacklining: A group of people have stretched a tightrope across a gym. They \textit{\textcolor{red}{drop an orange rope at the end.}}
\end{minipage}}\tabularnewline
\bottomrule
\end{tabular}
\caption{A Five-Shot Testing Data Example of the HellaSwag Task for the ALLM-AC Inference.}
\end{subtable}

\caption{Testing Data Examples for the AnLLM-EP and AnLLM-AC models. The log-likelihood of the red italicized texts is calculated as the choice probabilities.}
\label{tab:testingdata}

\end{table*}

\end{document}